%% file: egbib.tex
\documentclass[10pt,twocolumn,letterpaper]{article}

\usepackage{iccv}
\usepackage{times}
\usepackage{epsfig}
\usepackage{graphicx}
\usepackage{amsmath}
\usepackage{amssymb}
\usepackage{xcolor}
\usepackage{float}
\usepackage{nicefrac}

\usepackage[breaklinks=true,bookmarks=false]{hyperref}

\iccvfinalcopy 

\def\iccvPaperID{****} 

\ificcvfinal\fi

\begin{document}

\title{Is context all you need? Scaling Neural Sign Language Translation to Large Domains of Discourse}

\author{Ozge Mercanoglu Sincan$^1$, Necati Cihan Camgoz$^2$, Richard Bowden$^1$\\
$^1$University of Surrey, United Kingdom\\
$^2$Meta Reality Labs, Switzerland\\
{\tt\small \{o.mercanoglusincan, r.bowden\}@surrey.ac.uk, neccam@meta.com}
}

\maketitle
\ificcvfinal\thispagestyle{empty}\fi

\begin{abstract}
Sign Language Translation (SLT) is a challenging task that aims to generate spoken language sentences from sign language videos, both of which have different grammar and word/gloss order. From a Neural Machine Translation (NMT) perspective, the straightforward way of training translation models is to use sign language phrase-spoken language sentence pairs. However, human interpreters heavily rely on the context to understand the conveyed information, especially for sign language interpretation, where the vocabulary size may be significantly smaller than their spoken language equivalent.

Taking direct inspiration from how humans translate, we propose a novel multi-modal transformer architecture that tackles the translation task in a context-aware manner, as a human would. We use the context from previous sequences and confident predictions to disambiguate weaker visual cues. To achieve this we use complementary transformer encoders, namely: (1) A Video Encoder, that captures the low-level video features at the frame-level, (2) A Spotting Encoder, that models the recognized sign glosses in the video, and (3) A Context Encoder, which captures the context of the preceding sign sequences. We combine the information coming from these encoders in a final transformer decoder to generate spoken language translations.

We evaluate our approach on the recently published large-scale BOBSL dataset, which contains $\sim$1.2M sequences, and on the SRF dataset, which was part of the WMT-SLT 2022 challenge. We report significant improvements on state-of-the-art translation performance using contextual information, nearly doubling the reported BLEU-4 scores of baseline approaches. 
\end{abstract}

\input{1_intro.tex}

\input{2_relatedwork.tex}
\input{3_method.tex}
\input{4_experiments.tex}

\input{5_conclusion.tex}

\section*{Acknowledgement}
This work was supported by the EPSRC project ExTOL (EP/R03298X/1), SNSF project `SMILE II’ (CRSII5 193686), European Union’s Horizon2020 programme (`EASIER’ grant agreement 101016982) and the Innosuisse IICT Flagship (PFFS-21-47). This work reflects only the authors view and the Commission is not responsible for any use that may be made of the information it contains.

{\small
\bibliographystyle{ieee_fullname}
\bibliography{egbib}
}

\end{document}

%% file: 1_intro.tex
\section{Introduction}
\label{sec:intro}

\begin{figure}  
\centering
    \includegraphics[width=0.5\textwidth]{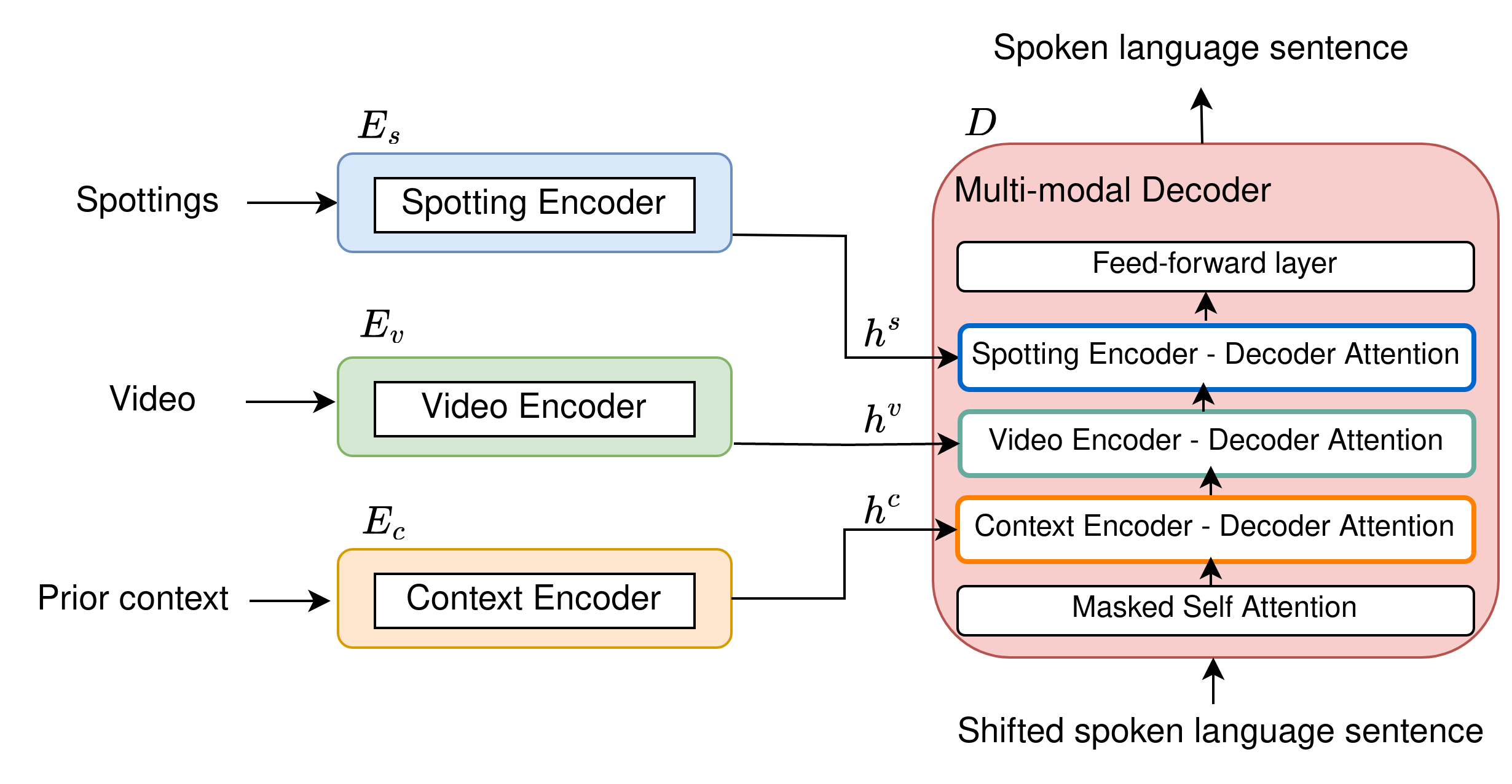}
    \caption{An overview of the proposed multi-modal sign language translation architecture. }
    \label{fig:network}
\end{figure}

Sign languages are visual languages and the primary languages of Deaf communities. They are languages in their own right, as rich as any spoken language, and can vary considerably between countries with strong dialect differences within a country \cite{emmorey2001language}. They have their own lexicons and grammatical constructs, thus converting between sign and spoken language is a translation problem. 

Sign Language Recognition (SLR) \cite{rastgoo2021sign} and Sign Language Translation (SLT) are active research areas within computer vision \cite{camgoz2018neural, chen2022simple, yin2022mlslt, zhou2021improving}. While SLR focuses on the recognition of signs within a video, SLT aims to generate meaningful spoken language interpretation of a given signed phrase or vice versa.
In our work, we focus on the former part of SLT, namely translating continuous sign videos into spoken language sentences.

Automatic SLT is a challenging problem for a number of reasons. Firstly, as stated, sign languages have their own grammar, they are not translated simply word-by-word by replacing words with signs \cite{sutton1999linguistics}. Secondly sign languages contain many channels that are used in combination i.e. hand articulation, facial expression, and body posture are all used in combination and their use can vary depending on context. For example, the hand shape of a sign may change depending on the context. A good example of this would be the verb `\emph{to give}'. The verb `\emph{give}' is directional and the direction of the motion is subject to the placement of objects and use of space in front of the signer. But the hand shape can also change depending on the type of object being given. Thirdly, motions can be subtle or fast and this leads to motion blur. Finally, many sign can also look very similar. All of these factors make it difficult to recognize the sign that is being performed without the context in which it is used. 

Human interpretation or translation of sign languages heavily relies on context, as it is fundamental to all language understanding. Consider the use of homophones in spoken language. An active listener has no issues in the disambiguation of homophones despite the fact there are no auditory cues to help. This is because we are able to use the context to disambiguate the meaning of the homophone. However, much of the SLT work to date has neglected such context, focusing largely on sentence pairs. In fact, most machine translation datasets shuffle the order of sentences, making it impossible to utilize the context from the previous sentences.

In this work, we propose a novel sign language translation architecture that incorporates important contextual information. It combines weak visual cues from a 3D convolutional backbone with strong cues from the context and sparse sign spottings. An overview of the approach can be seen in Figure \ref{fig:network}. 

We evaluate our approach on the largest available sign language dataset, BOBSL \cite{albanie2021bbc}, which covers a wide domain of various topics. 
We obtain significant performance improvements by incorporating context  and automatic spottings (1.27 vs. 2.88 in BLEU-4). We also evaluate our approach on the WMT-SLT 2022 challenge data, specifically the SRF partition, and surpass the reported performance of all challenge participants.

The contributions of this paper can be summarized as:
\begin{itemize}
    \item We propose a novel multi-modal transformer network that incorporates the context of the prior information and automatic spottings.
    \item We conduct extensive experiments to examine the effects of different approaches to capturing context.    
    \item Our approach achieves state-of-the-art translation performance on two datasets, namely BOBSL, the largest publicly available sign language translation dataset, and the WMT-SLT 2022 challenge data.
    
\end{itemize}

The remainder of the paper is organized as follows: In Section 2, we summarize the related work. In Section 3, we describe our proposed sign language translation network. In Section 4, we provide information about the datasets we use and provide model training details. Section 5 presents the experimental results of the proposed method and we conclude the paper in Section 6.

%% file: 2_relatedwork.tex
\section{Related Work}
\textbf{Sign Language Recognition (SLR)} has seen consistent research effort from the computer vision community for decades \cite{cooper2011sign, starner1998real, zafrulla2011american}. The advances in models and techniques, also the release of recent isolated \cite{albanie2021bbc, huang2018attention, li2020transferring, sincan2020autsl}, and continuous \cite{huang2018video, koller2015continuous} SLR datasets have led to significant improvements in the accuracy and robustness of sign language recognition systems.

SLR can be grouped into two sub-problem; isolated and continuous SLR. While the isolated SLR videos contain only a single sign, continuous SLR videos contain multiple sign sequences.  After the emergence of 2D convolutional neural networks (CNNs), 2D CNNs were quickly applied to model the visual appearance in SLR \cite{neverova2015moddrop, pigou2014sign, pigou2018beyond, sincan2020autsl}. Sequence models such as the recurrent neural network (RNN) \cite{pigou2018beyond}, long short-term memory (LSTM) \cite{sincan2020autsl}, hidden markov model (HMM) \cite{tur2021evaluation} have all been used to encode temporal information. Following 2D CNNs, 3D CNNs were developed and have achieved state-of-the-art performance on a wide range of computer vision tasks, including sign language recognition \cite{albanie2020bsl, huang2018attention, joze2018ms, li2020transferring}. 

In addition to images, researchers have also used other input modalities for SLR, such as depth, skeleton, optical flow, and motion history image (MHI) to improve recognition accuracy \cite{hu2021signbert, jiang2021skeleton, neverova2015moddrop, sincan2020autsl, sincan2022using}. Some studies also introduced the use of different cues such as cropped hands and faces \cite{cihan2017subunets, gruber2021mutual}, or an attention mechanism \cite{de2021isolated, sincan2022using} to obtain better discriminative features.

These advances in the field of isolated SLR have also been applied to continuous SLR. Since continuous SLR videos contain multiple co-articulated signs, it is a more challenging problem. The explicit alignment between the video sequence and gloss sequence generally does not exist. In order to tackle this problem, Connectionist Temporal Classification (CTC) \cite{graves2006connectionist} is widely used \cite{cihan2017subunets, hao2021self, pu2019iterative, zuo2022c2slr}.

\textbf{Sign Language Translation (SLT)} is still in its infancy due to the lack of large-scale sign language translation datasets. While machine translation datasets for spoken languages contain many millions of sentence pairs such as 22.5M for English-French, and 4.5M for English-German pairs (WMT shared tasks \cite{bojar2014findings}), the first public SLT dataset PHOENIX14-T \cite{camgoz2018neural}, which was released in 2018, had only 8K sentences and its domain of discourse was limited to weather forecast. The authors handle the SLT as a Neural Machine Translation (NMT) problem and proposed the first end-to-end SLT model by combining CNNs with the attention-based encoder-decoder network with RNNs.

One of the most significant advances in NMT was the introduction of the Transformer network by Vaswani et. al \cite{vaswani2017attention}, which is based solely on attention mechanisms and waives recurrent networks, for the sequence transduction problem. Camgoz et al. \cite{camgoz2020sign} applied transformer architecture to the sign language translation problem. In recent years, transformers have become popular in SLT \cite{dey2022clean, yin2022mlslt, yin2020better, zhou2021improving}. Some studies tackle SLT with a two-stage approach, i.e., in the first part glosses are recognized from sign videos (Sign2Gloss), and then glosses are mapped into a spoken language sentence (Gloss2Text) \cite{camgoz2018neural, yin2020better}. 
On the other hand, some studies deal with an end-to-end solution that predicts the spoken language sentence from sign video inputs \cite{camgoz2020sign, zhou2021improving}. 

Zhou et al. \cite{zhou2021improving} proposed a two-stage approach, but unlike others, their approach is based on back-translation. They convert spoken language text to sign sequences with both text-to-gloss and gloss-to-sign steps to generate synthetic data. They used the synthetic samples as additional data and trained an end-to-end SLT method based on the transformer. Zhou et. al \cite{zhou2021spatial} and Camgoz et. al \cite{camgoz2020multi} also utilized multiple cues for the SLT task, such as hands and face. To the best of our knowledge, and perhaps surprisingly, using context has not been exploited in the literature. However, Papastratis et. al \cite{papastratis2021continuous} did use the previous sentence to initialize the hidden state of a BLSTM for predictions of the next video sequence to improve recognition accuracy in a continuous SLR. They obtain slightly better results when the context-aware gloss predictions were fed into the transformer for SLT.

\textbf{Datasets:} PHOENIX14-T \cite{camgoz2018neural} became the most commonly used dataset in the literature. The performance on this dataset is generally satisfactory to provide a usable translation, e.g., Chen et. al \cite{chen2022simple} obtained 28.39 in terms of BLEU-4 score. However, due to its limited domain of discourse, models trained on PHOENIX14-T have little real-world applicability. To address this, researchers released several datasets in recent years \cite{albanie2021bbc, camgoz2021content4all, mullerfindings}. The largest to date is BOBSL \cite{albanie2021bbc}, a broadcast interpretation-based large-scale British Sign Language (BSL) dataset. Their SLT baseline is based on the transformer network and obtains only 1.0 in terms of BLEU-4.
Recently, Swiss German Sign Language (DSGS) broadcast datasets were introduced in the first SLT-WMT shared task \cite{mullerfindings}, where all the submissions scored under 0.56 in terms of BLEU-4.
 Yin et. al \cite{yin2022mlslt} collect the first multi-lingual dataset for multiple sign language translations and proposed the first multi-lingual SLT model. Although significant progress has been made in the area of SLT, there is still room for further improvement.

%% file: 3_method.tex
\section{Method}

Most sign translation datasets and especially those based on broadcast interpretation \cite{albanie2021bbc, camgoz2018neural, mullerfindings}, contain a set of consecutive sign phrase videos $(V_1, ...., V_M)$ and spoken language sentences $(S_1, ..., S_N)$. In some datasets, such as Phoenix2014T \cite{camgoz2018neural}, sign phrase videos and their spoken language translations are paired and the order of the pairs are shuffled and distributed between training and evaluation sets. Unfortunately, this destroys the context of the sentence.  Datasets like BOBSL \cite{albanie2021bbc} release the video and sentence sets with only weak alignment. Although this is generally regarded as a weakness, making subsequent learning from the data more challenging, it has a fundamental advantage that we make use of in this work: it allows the use of context to improve the translation. 

Given an input video $V = (x_1, x_2, ..., x_T)$ with $T$ frames, the aim of a sign language translation is to learn the conditional probability $p(S|V)$ in order to generate a spoken language sequence $S = (w_1, w_2, ..., w_U)$ with $U$ words.

We propose to take advantage of the contextual information that comes from the preceding context, $S_C = (S_{n-1}, S_{n-2}, S_{n-3}, ...)$. We also make use of sparse sign spottings, $Sp = (g_1, ..., g_K)$, automatically recognized from the current video $V$ using a state-of-the-art model. Thus, we extend the classical translation formalization to one of learning the conditional probability $p(S|V, S_C, Sp)$. This conditioning allows weak and ambiguous visual cues in $V$ to be disambiguated based on context.

\begin{figure*}  [htbp] 
    \centering
    	\includegraphics[width=0.85\textwidth]{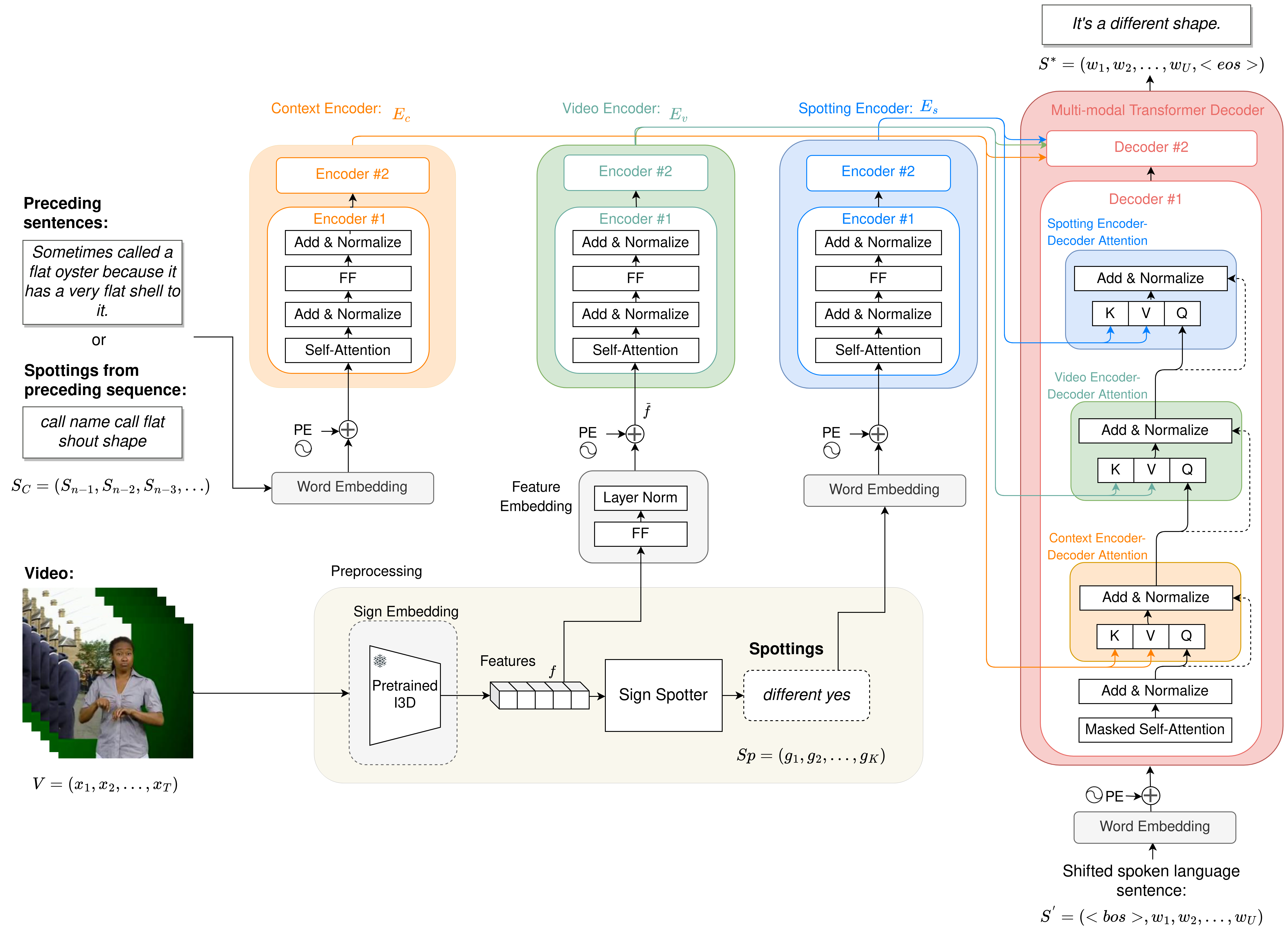}
        \caption{A detailed overview of the proposed multi-modal sign language translation architecture. }
    	\label{fig:networkDetailed}
    \end{figure*}

Our translation network is based on a transformer architecture and contains three separate encoders, $E_v, E_c, E_s$ for each of the different input cues, i.e., video, context, and spottings, and a multimodal decoder, $D$, which learns the mapping between all input source representations and the target spoken language sentence. A detailed overview of our model is shown in Figure \ref{fig:networkDetailed}.

\subsection{Embedding Layers}

Following the classic neural machine translation methods, we first project source and target sequences to a dense continuous space via embedding layers. In order to represent video sequences, we utilize pretrained CNNs. For linguistic concepts that originate from written text in the form of the preceding and target spoken language sentences and spotted sign glosses, we use word embedding layers. 

\textbf{Sign Embedding:} 
To convert a given video, $V$, to its feature representation, we use the I3D model \cite{carreira2017quo} as a backbone due to its recent success in sign recognition tasks. We first divide the videos into smaller video clips, $c_t = (x_t, ..., x_{t+L-1})$ of size $L$. In our experiments we use a window size of $L=16$ to obtain the sign video embedding:

\begin{equation}
    f_t = \mathrm{SignEmbedding}(c_t)
\end{equation}
We stride $SignEmbedding$ over the full video $V$ with the step size of $4$, thus yielding a final feature set of $f_{1:\frac{T-L}{4}+1}$. 

We considered two types of features as the output of our sign embedding layer, namely a) 1024-dimensional representation that is extracted from the last layer before classification, and b) C-dimensional class probabilities after the softmax activation function. We conduct experiments using both of these feature representations in our translation pipeline and compare their performance.

\textbf{Feature Embedding:} To avoid biases caused by dimensionality we project the extracted feature representations into the same size denser space using a linear layer. We also employ layer normalization to transform them to be of the same scale. This feature embedding operation can be formalized as:
\begin{equation}
    \hat{f}_t = \mathrm{FeatureEmbedding}(f_t)
\end{equation}

\textbf{Word Embedding:} 
We first tokenize our spoken language sentences using a pretrained BERT model \cite{devlin2018bert}. More specifically we employ the \emph{BERT-base-cased} and \emph{BERT-base-german-cased} from the Huggingface's Transformer library \cite{wolf2019huggingface}, which uses WordPiece tokenization. The word embedding layer is shared between all the text input cues, such as spottings, context sentences, and shifted target sentences.

\textbf{Positional Encoding:} 
In order to provide sequential order information to our networks we use the standard positional encoding method as proposed in \cite{vaswani2017attention} in the form of shifted sine and cosine waves. This is added after the feature and word embedding layers. This positional encoding can be formalized as:
\begin{equation}
    \bar{f}_t = \mathrm{PositionalEncoding}(\hat{f}_t)
\end{equation}

\subsection{Translation Network}

After embedding layers, positionally encoded features and word vectors are sent to the transformer encoders. Our encoders have a stack of two identical layers each of which has a multi-headed self-attention and a fully connected feed-forward layer. Each of these two sub-layers is followed by a residual connection and layer normalization. 

\textbf{Video-Encoder:} The video encoder network, $E_v$, takes the positionally encoded feature vectors $\bar{f}_{1:\frac{T-L}{4}+1}$ that come from the feature embedding layer and produces a spatial-temporal representation $h_{1:\frac{T-L}{4}+1}^v$ that captures the motion and content of the video. 

\textbf{Context-Encoder:} The context encoder, $E_c$, takes positionally encoded-word embedding results from preceding context, $S_C$, and produces representations, $h^c$, which capture the context in which the currently signed phrase is performed.

\textbf{Spotting-Encoder:} The spotting encoder network, $E_s$, takes positionally encoded spotting embeddings $Sp$ and produces representations $h^s$ that correspond to confident but sparse sign detections that have been spotted in the current video, $V$, that we are attempting to translate.
See Section \ref{signspotter} for details of the sign spotting technique \cite{momeni2022automatic} we used in our experiments.

\textbf{Decoder:} After encoding each input modality, the output of the encoder layers $h^c, h^v, h^s$, and the positionally encoded and shifted spoken sentence is then sent to the transformer decoder $D$. We extend the classical transformer decoder architecture \cite{vaswani2017attention} by introducing several encoder-decoder attention layers, which combine and enrich the representations coming from complementary cues of information. The network flow can be formalized as:
\begin{equation}
    \begin{array}{l}
        h^c = \mathrm{ContextEncoder}(S_C) \\
        h^v = \mathrm{VideoEncoder}(V) \\
        h^s = \mathrm{SpottingEncoder}(Sp) \\
        S^* = \mathrm{Decoder}(h^c, h^v, h^s, S') \\
    \end{array}
\end{equation}

where $S'$ and $S^*$ correspond to the shifted and predicted target sentences, respectively. In words, firstly, the word embeddings extracted from the shifted spoken language embedding $S'$ are passed to the masked self-attention layer. Then, our first encoder-decoder attention layer takes outputs of the masked self-attention and context encoder, $h^c$. The output of the context encoder-decoder attention is sent to the video encoder-decoder attention to be used as a query, while the key and the value come from the video encoder, $h^v$. In a similar way, the spotting encoder-decoder attention performs attention operations over $h^s$ and the previous layer. Finally, the last representation of the transformer decoder is projected to the space of the target vocabulary using a linear layer to predict the target spoken language sentence $S^*$, one word at a time. 

We train our network using cross-entropy loss as proposed in \cite{vaswani2017attention}, by comparing the predicted target sentence $S^*$ against the ground truth sentence $S$ at the word level.

%% file: 4_experiments.tex
\section{Dataset and Implementation Details}
\subsection{Datasets}

\textbf{SRF} is a Swiss German Sign Language (DSGS) dataset that was recently released for the WMT-SLT 2022 challenge \cite{mullerfindings} as one of the training corpora. It contains daily news and weather forecast broadcast. It includes 16 hours of sign footage, divided into 29 episodes, performed by three signers. In total 7,071 subtitles were manually aligned by Deaf annotators. Separate development and test sets were provided in the WMT-SLT. We use the SRF dataset for training and used the official development and test sets for the evaluation of the model to be able to compare our approach against the methods presented in the challenge.

\textbf{BOBSL} \cite{albanie2021bbc} is a large-scale British Sign Language (BSL) dataset that consists of BSL-interpreted BBC broadcast footage covering a wide range of topics. The dataset has an approximate duration of 1,400 hours and contains around 1.2M sentences. While the training and validation set's subtitles are audio-aligned, the test data is manually aligned and contains 20,870 sentences with  a vocabulary size of 13,641.

\subsection{Sign Spotter}
\label{signspotter}
Momeni et al. \cite{momeni2022automatic} released automatically extracted dense annotations for the BOBSL dataset. We use these annotations as the spotting input on the BOBSL experiments.

The key idea is that a set of video clips with a particular sign must have a correlation at the time when the sign is performed. Taking inspiration from \cite{momeni2022automatic}, we create similar automatic dense annotations for the SRF dataset by correlating the I3D features and examplar subtitles. To do this, firstly we lemmatize and lowercase each word in the subtitle sentences and extract a vocabulary list. German language has compound words by concatenating two words. In order to reduce the number of singletons in the vocabulary list, we use the compound-split library \cite{compound-split}. Then, for each word $w$, we take a reference video clip $V_0$ that contains $w$ in its subtitle sentence. We choose random $N=9$ positive video examples $V_1, V_2, .., V_N$ that contain the word $w$ in their subtitles, and $3*N$ negative video examples that do not contain $w$ to avoid annotating non-lexical signs. We compute the cosine similarities between reference and examplar video features. We apply a voting scheme among the videos with cosine similarity above 0.5 to find the location of the given word in the reference video.

\subsection{Implementation Details}

\textbf{Sign Embeddings:} For full-body video inputs, we pretrain two different I3D models which we call BSL-I3D and DGS-I3D on two different sign language datasets, namely BOBSL \cite{albanie2021bbc} and MeineDGS \cite{konrad2019meine}. While training the BSL-I3D model we use the annotations released with the dataset \cite{albanie2021bbc} which has a vocabulary size of 2,281. For MeineDGS we use the linguistic annotation available with the dataset. In order to obtain a similar size vocabulary of 2,301 classes, we choose classes that have more than 12 occurrences.
    
We resize the input images to $224\times224$ and follow the training instructions of \cite{albanie2021bbc} with some small modifications; we use the Swish activation function instead of ReLU and change the learning rate scheduler to reduce on a plateau. We also use label smoothing of 0.1 in order to help reduce overfitting.

\textbf{Training and Network Details:} Our model is implemented using PyTorch \cite{paszke2019pytorch}. We use the Adam \cite{kingma2014adam} optimizer with an initial learning rate of $3\times10^{-4}$ ($\beta_1$ = 0.9, $\beta_2$ = 0.999, $\epsilon$ = $10^{-8}$ ) with a batch size 16 on SRF; and learning rate $6\times10^{-4}$ with batch size 64 on the BOBSL dataset. We reduce the learning rate by a factor of 0.7, if the BLEU-4 score does not increase for five epochs. This step continues until the learning rate drops below $10^{-5}$.

For transformer encoders and the decoder, we use two layers with 8 heads. We conduct an ablation study to choose the size of the hidden layers and the feed-forward layers (in section 5.1). We choose 512 and 1024, respectively. We use 0.1 for the dropout rate.

During training, we use a greedy search to evaluate translation performance on the development set. At inference, we evaluated both a greedy search and a beam search (decoding size of 2 and 3) for our video-to-text approach. However, we did not observe a significant improvement in scores. Therefore, we provide greedy search performances on both validation and test set.

\textbf{Metrics:} We use BLEU-1, BLEU-4 \cite{papineni2002bleu}, ROUGE \cite{lin2004rouge}, and CHRF \cite{popovic2015chrf} scores, which are commonly used metrics for machine translation, to evaluate the performance of our model. As ROUGE, we use ROUGE-L F1 score; as BLEU score we use the sacreBLUE \cite{post-2018-call} implementation.

\section{Experiment Results}

We run our experiments in an end-to-end manner on two recently released sign language datasets, namely the SRF partition of WMT-SLT \cite{mullerfindings} and BOBSL \cite{albanie2021bbc},  which is the largest sign language dataset available.
For each dataset, we train baseline models that have one encoder and one decoder, and take only one input source, i.e., the preceding context (using the preceding spoken sentence or preceding spottings), current spotting, or video. We name our single modality models as \textit{Context-to-Text, Spot-to-Text, Video-to-Text}, respectively.

Then, we investigate the impact of integrating context information to the \textit{Spot-to-Text} or \textit{Video-to-Text} approaches by adding a context encoder and using a dual-mode transformer decoder with the related encoder-decoder attention layers. 
Finally, we investigate using all sources simultaneously to gain more information. We combine all three sources using three separate encoders and a decoder. We name our final model as \textit{Context+Video+Spot-to-Text}.

\subsection{Experiments on SRF partition of WMT-SLT} 
 
\textbf{Video-to-Text:} We evaluate our \textit{Video-to-Text} model which takes only the video source and tries to generate spoken language in an end-to-end manner.

First, we conduct ablations studies on the SRF partition of the WMT-SLT dataset using different types of input channels for the \textit{Video-to-Text} model. We run our experiments with different numbers of hidden size (HS) and feed-forward (FF) units, with $64\times128$, $128\times256$, $256\times512$, $512\times1024$, $512\times2048$. We obtain similar results with $512\times1024$ and $512\times2048$, where $512\times1024$ is slightly better. Therefore, for the rest of our experiments, we use $512\times1024$ parameters for HS$\times$FF. 

Table \ref{tab:singleChSrf} shows our ablation experiments against the baseline \cite{mueller2022sign-sockeye-baselines} on the WMT-SLT development set. We repeat each experiment 3 times and report the mean and standard deviation of scores. All our experiments outperform the baseline. 

We do not observe any significant difference between the BSL or DGS-pretrained I3D model on the WMT-SLT. Our best score, obtained using BSL-I3D features, was 1.51 in terms of BLEU-4. On the other hand, class probabilities obtain lower BLEU scores than feature embeddings. Therefore we use BSL-I3D features going forward for our video encoder.

\begin{table}[!ht]
	\centering
 \resizebox{\linewidth}{!}{
	\begin{tabular}{ lc|ccc}
		\hline
		\textbf{} & \textbf{Size} &\textbf{BLEU-1} & \textbf{BLEU-4}  &  \textbf{CHRF}    \\ \hline
        Baseline \cite{mueller2022sign-sockeye-baselines} & & - & 0.58  & -  \\\hline
         BSL-P  & 2281 & 14.26 ± 0.47 &1.01	± 0.2  & 17.0 ± 0.17  \\
         DGS-P  & 2301 &  14.6 ± 0.55  &1.03	± 0.08 & 17.03	± 0.47 \\ 
         BSL-F  & 1024 &	15.86 ±  0.2  & 1.23 ± 0.25 & 17.27	± 0.15 \\
         DGS-F  & 1024 & 15.14 ± 0.44  &1.17± 0.08 & 17.13	± 0.12 \\
         \hline
	\end{tabular}} 
    \caption{Evaluation of different features for SLT on WMT-SLT development set. BSL-F: BSL-I3D features, DGS-F: DGS-I3D features, BSL-P: BSL-I3D class probabilities, DGS-P: DGS-I3D class probabilities.}
    \label{tab:singleChSrf}
	\end{table}

    \begin{table}[!ht]
	\centering
 \resizebox{\linewidth}{!}{
	\begin{tabular}{ l|cccc}
		\hline   
		\textbf{} & \textbf{BLEU-1} & \textbf{BLEU-4}  &  \textbf{CHRF} & \textbf{ROUGE}    \\ \hline
        
         MSMUNICH \cite{dey2022clean} & - & 0.56  & 17.4 & -  \\
         SLT-UPC  \cite{tarres2022tackling}  & - & 0.5  & 12.3 & -  \\
         SLATTIC \cite{shi2022ttic} & - & 0.25 & 19.2 & - \\ 
         Baseline  \cite{mullerfindings} & - & 0.12  & 5.5 & - \\ 
         DFKI-MLT \cite{hamidullah2022dfki} & - & 0.11 & 6.8 & - \\
         NJUPT-MTT & - & 0.10 & 14.6 & -\\
         DFKI-SLT \cite{hufe2022experimental} & - & 0.08 & 18.2 & -\\ \hline
          Ours   &   & \\
          - Video-to-Text & 14.43 & 0.81 & 18.18 & 5.60 \\
          - Context-to-Text  & 12.80 & 0.69 & 14.48 & 3.73 \\ 
         - Context+Video-to-Text & 14.33 & 1.00 & 18.12 & 6.00 \\ \hline
         - Spot-to-Text & 22.11 & 1.87 & 22.23 & 11.17 \\
         
           - Context+Video+Spot-to-Text &  31.36 & 3.93 & 24.69 & 17.65\\ \hline
	\end{tabular}
 }
 	\caption{Comparison with the literature on the full WMT-SLT test set.}
  \label{tab:srf2} 
	\end{table}

Table \ref{tab:srf2} shows the comparison of our approach against the participants of the WMT-SLT shared task \cite{mullerfindings}. All approaches are based on Transformer architectures \cite{vaswani2017attention}. Similar to our \textit{Video-to-Text}, MSMUNICH \cite{dey2022clean} also uses an I3D model for feature extraction and obtained the highest score of 0.56 in BLEU-4. While they use an I3D model pretrained on BSL-1K \cite{albanie2020bsl}, we pretrained our I3D on the BOBSL \cite{albanie2021bbc} which provides better feature representation and a slight improvement.

 \begin{table*}[!ht]
	\centering
    \small
	\begin{tabular}{ l|cccc|cccc}
		\hline
        \textbf{}  & \multicolumn{4}{c}{\textbf{Val}}  & \multicolumn{4}{c}{\textbf{Test}} \\  
        
		\textbf{}  & \textbf{BLEU-1}  &  \textbf{BLEU-4} &\textbf{ROUGE} &\textbf{CHRF}  & \textbf{BLEU-1}  &  \textbf{BLEU-4} &\textbf{ROUGE} &\textbf{CHRF} \\ \hline
        \textbf{Context-to-Text }   & & & & & & & &   \\
         - 1 preceding sentence & 13.50 & 0.45 & 5.59 & 10.0 & 13.41 & 0.45 & 6.10 & 10.4  \\
         - 2 preceding sentence & 13.21 & 0.52 & 5.11 & 10.2 & 13.34 & 0.42 & 5.54 & 10.4 \\
         - 3 preceding sentence & 13.32 & 0.51 & 5.36 & 10.2 & 13.0 & 0.43 & 5.59 & 10.5\\  
         - Max 10 spottings & 13.77 & 0.74 & 6.33 & 10.88 & 12.90 & 0.60 & 6.01  & 10.76  \\
         - Max 20 spottings  & 13.88 & 0.75 & 6.36 & 10.9 & 12.96 & 0.56 & 6.07 &  10.66 \\
        
        \hline
        
        \textbf{Spot-to-Text}   & 21.97 & 2.25 & 8.52 & 19.4 & 21.63 & 2.21 & 9.45 & 19.7\\ \hline

        \textbf{Context+Spot-to-Text}   & 22.77 & 2.56 & 9.98 & 19.9 & 21.68 & 2.43 & 10.0 & 19.72 \\ \hline
              
	\end{tabular}
 	\caption{Performance of our text-to-text models on the BOBSL dataset.}
    \label{tab:Bobsl1}
	\end{table*}

  \begin{table*}[!ht]
   \centering
   \small
	\begin{tabular}{ l|cccc|cccc}
		\hline
        \textbf{} & \multicolumn{4}{c}{\textbf{Val}}  & \multicolumn{4}{c}{\textbf{Test}} \\  
		\textbf{}   & \textbf{BLEU-1}  &  \textbf{BLEU-4} &\textbf{ROUGE} &\textbf{CHRF}  & \textbf{BLEU-1}  &  \textbf{BLEU-4} &\textbf{ROUGE} &\textbf{CHRF} \\ \hline
        Albenie et. al \cite{albanie2021bbc}  & -&-&-& -& 12.78 & 1.00 & 10.16 & -\\ \hline
        
        \textbf{Video-to-Text} & & & & & & & & \\ 
        - trained with 274K & 15.15 & 1.02 & 12.71 & 19.7 & 12.68 & 0.83 & 8.32 & 17.9 \\ 
       - trained with 1M & 18.8 &  1.28 & 7.91 & 17.7 & 17.71 & 1.27 & 8.9 & 18.8\\ 
        \hline
        
        \textbf{Context+Video-to-Text }  & & & & & & & &  \\
        - 1 preceding sentence   & 20.18  &  1.53 & 9.13 & 18.2 & 19.11 & 1.51 & 9.94 & 19.3\\
        - 2 preceding sentence   & 19.14 & 1.52 & 8.97 & 18.0 & 18.15 & 1.41 & 9.56 & 18.9\\
        - 3 preceding sentence  & 20.05 & 1.56 & 9.08 & 18.1 & 18.82 & 1.48 & 9.64  & 19.1 \\ 
        - Max 10 spottings & 20.84 & 1.71 & 10.03 & 18.21 & 19.05 & 1.50 & 9.95 & 18.94 \\
        \hline

        \textbf{Context+Video+Spot-to-Text}  & & & & & & & &  \\
        -with 1 preceding sentence &  25.06 & 2.73 & 11.12 & 22.6  & 24.07 & 2.81 & 12.07 & 23.7\\   
        -with max 10 spottings & 25.94 & 3.07 & 12.27 & 23.69  & 24.29 & 2.88 & 12.41 & 24.53\\
        \hline
	\end{tabular}

    \caption{Impact of the integrating context and spottings information to video-to-text approaches on the BOBSL dataset.}
	\label{tab:Bobsl2}
	\end{table*}

\textbf{Context-to-Text:}  Here, we are testing how well a network can guess the content of a sentence given the context of the preceding sentence. To do this, we need ordered data. Although the development and test data of the SRF partition of WMT-SLT consists of segments extracted from several episodes, the segments contain consecutive numbers for each episode. Therefore, we used sorted segments to evaluate our \textit{Context-to-Text} model.  As can be seen in Table \ref{tab:srf2}, \textit{Context-to-Text}, which takes only the previous sentence as a source, performs worse than our \textit{Video-to-Text}. However, its BLEU-4 performance is still superior, and CHRF performance is competitive to the literature, which verifies that contextual information provides important cues for translation tasks.

\textbf{Context+Video-to-Text:} Next, we combine context and video sources by including a context encoder, a video encoder, and a decoder, which we call \textit{Context+Video-to-Text}. Incorporating context information besides video features improved our translation results as we expected.

 \textbf{Spot-to-Text:} In the literature, ground truth sign glosses are used to train a text-to-text translation model to create an upper bound for end-to-end translation \cite{camgoz2018neural}. Motivated by this, we created spottings as described in 4.2 using our BSL-I3D model. The trained \textit{Spot-to-Text} model achieves significantly better translation performance compared to other single-modality architectures.
 
\textbf{Context+Video+Spot-to-Text:} 
Finally, we integrate automatically created spottings as input to the spotting encoder. The performance gain is significant when compared to \textit{Context+Video-to-Text} and \textit{Spot-to-Text}, showing the benefits of the incorporation of complementary information cues.
However, this result should be taken as an upper bound on performance as the spotting approach requires a prior over the spoken word occurrence. This artificially inflates the performance but as can be seen, the potential benefits of accurate spotting on translation are significant.

\subsection{Experiments on BOBSL} 

\textbf{Context-to-Text:}  
 We evaluate training \textit{Context-to-Text} with two different types of data on the BOBSL; a) preceding sentences and b) preceding spottings. 
 As can be seen in Table~\ref{tab:Bobsl1}, using only the preceding text data leads to poor translation. Firstly, we experiment with increasing the context by providing more preceding text. While using more sentences provides a slight improvement in terms of the BLEU-4 and CHRF scores on the validation set, it did not help in the other scores or on the test set. In the experiments with preceding spottings, we experiment with different numbers of spottings. We use the spottings from up to 3 previous sentences since we do not see a significant improvement when we include more prior sentences in the previous experiments. 
 We obtain better results when we use the spottings of 3 prior sentences, but limit the maximum number of spottings to just 10.

\textbf{Spot-to-Text:} 
We utilized the sign spottings \cite{momeni2022automatic} of the BOBSL to evaluate our \textit{Spot-to-Text} model. We train our model using all automatic annotations without any thresholding, which obtains 21.63 and 2.21 for BLEU-1 and BLEU-4 on the test set, respectively.

\textbf{Video-to-Text:}  
In \cite{albanie2021bbc}, the authors provide an SLT baseline that is trained on a subset of the BOBSL training set. They created their new training set for sign language translation by filtering the sentences that contain high-confidence automatic spottings. They selected words that occur at least 50 times in the training set and constructed sentences by filtering according to this vocabulary. They also discard sentences with over 30 words, yielding 274K sentences. To provide a comparison, we first train our \textit{Video-to-Text} network on this subset. However, transformers tend to get better results with more data.
Therefore, we also train our model using all sentences as in the \textit{``version v1\_2"} of the BOBSL dataset for which the training set contains about 1M sentences. In this experiment, our BLEU-4 score increased to 1.27 from 0.83 as seen in Table \ref{tab:Bobsl2}. 

\textbf{Context+Spot-to-Text:} Firstly, we combine context and spotting sources by having a context encoder, a spotting encoder, and a decoder, which we call \textit{Context+Spot-to-Text}. We set the maximum number of spottings to 10. \textit{Context+Spot-to-Text} achieved better results than \textit{Spot-to-Text} (2.21 vs 2.43 BLEU-4 score in the test set).

\textbf{Context+Video-to-Text:} Then, we evaluate the \textit{Context+Video-to-Text} model. We use all training videos and all validation videos in our multi-modal experiments. As can be seen in Table \ref{tab:Bobsl2}, when using prior spoken text as input for context-encoder, our \textit{Context+Video-to-Text} model achieves a significant improvement over our \textit{Video-to-Text} model on both the manually aligned test set (1.27 vs. 1.51 BLEU-4, and 12.68 vs. 19.11 BLEU-1) and validation set (1.28 vs. 1.53 BLEU-4, and 18.8 vs. 20.18 BLEU-1). We also investigate using a different number of preceding sentences. Similar to \textit{Context-to-Text} experiments, increasing the number of preceding sentences does not improve the translation quality. On the other hand, we experiment with the preceding spottings for the context-encoder. Although we obtain a much better result in the validation set (1.53 vs. 1.71 BLEU-4), we get similar results in the test set. This shows that using either the preceding sentences or preceding spottings provides similar context and helps to provide better translation.

\textbf{Context+Video+Spot-to-Text :} Finally, we train our transformer using all modalities. Our final approach is able to surpass all previous models and obtains state-of-the-art on the BOBSL dataset test set, with 2.81 for BLEU-4 and 24.07 for BLEU-1.

\textbf{Qualitative results:}  In this section, we share translations produced by the proposed model using different modalities and discuss our qualitative findings. As shown in Table \ref{tab:examples}, we compare our \textit{Video-to-Text}, \textit{Context+Video-to-Text} and \textit{Context+Video+Spot-to-Text} to better analyze the contribution of using the preceding context and current spottings. The results show that although translation quality is not perfect, context information helps us to get closer to the true meaning when compared to \textit{Video-to-Text}. As shown in the first example, the ground truth translation is \textit{``He lost nearly 200 sheep during the prolonged heavy snow in April."}. While \textit{Video-to-Text} model is able to infer only \textit{``sheep"} correctly, \textit{Context+Video-to-Text} model produces \textit{``Two sheep have been killed by the weather."}, which is a closer meaning.

\begin{table}[ht]
	\centering
 \def\arraystretch{1.3}
  \resizebox{\linewidth}{!}{
	\begin{tabular}{p{1.8cm}|p{6cm}}
		\hline
        \textbf{Ex\#1} GT: &  He lost nearly 200 sheep during the prolonged heavy snow in April.  \\
        V2T: & The sheep are rounded up and the autumn begins to drift away. \\
        (C+V)2T:  & The two sheep have been killed, and the two have been killed by the weather. \\
        (C+V+S)2T  & And the sheep are in the middle of April, and they're all farmed in the winter. \\ \hline
        
        \textbf{Ex\#2} GT: & You can see it's quite a different shape... \\
         V2T: & It's a very different story. \\
       (C+V)2T:  &  It's a very different shape. \\
        (C+V+S)2T  & It's a different shape. \\ \hline

        \textbf{Ex\#3} GT: & With the crops on the farm, summer is a busy time of year with harvest just around the corner. \\
         V2T: & It's a real dramatic change in the night and it's a real labour of love. \\
        C+V2T: & It's a very busy time of year, but it's a very busy time of year. \\
        (C+V+S)2T: & During the summer, the farm is busy grazing and the farm is busy harvesting. \\ \hline

         \hline
	\end{tabular}}
 	\caption{Qualitative results of the proposed method on the BOBSL. V2T: \textit{Video-to-Text}, (C+V)2T: \textit{Context+Video-to-Text}, (C+V+S)2T : \textit{Context+Video+Spot-to-Text}.}
    \label{tab:examples}
	\end{table}

%% file: 5_conclusion.tex
\section{Conclusion}

In this paper, we have proposed a novel multi-modal transformer architecture for context-aware sign language translation. Our approach utilizes complementary transformer encoders, including a spotting and video encoder for modeling the current sign phrase and a context encoder for capturing the context of preceding sign sequences. These encoders are then combined in a final transformer decoder to generate spoken language translations. We evaluate our approach on two sign language datasets with large domains of discourse and obtain state-of-the-art results by doubling the BLEU-4 score. We hope this work will encourage the exploration of new model ideas on large-scale sign language translation. A future direction may include exploring the leverage of context, such as to alleviate the local ambiguity for similar signs, or to improve spottings performance.